\documentclass[times,twocolumn,final,authoryear]{elsarticle}
			\usepackage{amsmath,amsfonts,amssymb}
			\usepackage{graphicx}
			\usepackage{mathtools}
			\usepackage{setspace}
		
			\usepackage{booktabs}

			\newcommand{\sq}{\vspace{-.2cm}}
			\usepackage{prletters}
			\usepackage{framed,multirow}
			\usepackage{fancyhdr}
			\usepackage{amssymb}
			\usepackage{latexsym}
			 \setcitestyle{square}
			 \usepackage[mathscr]{euscript}
			
			\usepackage{url}
			\usepackage{xcolor}
			\definecolor{newcolor}{rgb}{.8,.349,.1}

			\journal{Pattern Recognition Letters}
				
\begin{document}

			\begin{frontmatter}

			\title{Novel Framework for Spectral Clustering using 
			Topological Node Features(TNF)}
			
			\author[1]{Lalith Srikanth 
			\snm{Chintalapati}} 
		
			\author[2]{Raghunatha Sarma \snm{Rachakonda}\corref{cor1}}
			\cortext[cor1]{Corresponding author: }
			\ead{rraghunathasarma@sssihl.edu.in}
			
			\address[1,2]{Sri Sathya Sai Institute of Higher 
			Learning, Vidyagiri, 
			Prashanthi Nilayam, Anantpur Dist, Andhra Pradesh, 
			India. 515134}
			%\address[2]{Sri Sathya Sai Institute of Higher 
			%Learning, Vidyagiri, 
			%Prashanthi Nilayam, Anantpur and 515134, India}
			
			\received{1 May 2016}
			\finalform{10 May 2016}
			\accepted{13 May 2016}
			\availableonline{15 May 2016}
			\communicated{}

			\begin{abstract}
			%	\doublespacing 
			%Clustering is one of the important tasks in many image 	processing and pattern recognition applications. Among clustering techniques, 
			Spectral clustering has	gained 
			importance in recent years due to its ability to cluster complex data as it requires only pairwise similarity among data points with its ease of implementation. The central point in spectral clustering is the process of capturing pair-wise similarity. In the literature, many research techniques	have been proposed for effective construction of affinity 	matrix with suitable pairwise similarity. In this paper a general framework for capturing pairwise affinity using local features such as density, proximity and structural similarity is been proposed. Topological Node Features are exploited to define the notion of density and  local structure. These local features 	are  incorporated into the construction of the affinity matrix. Experimental results, on widely used datasets such as 
			synthetic shape datasets, UCI real datasets and MNIST handwritten datasets show that the proposed framework outperforms standard spectral clustering methods. 	
			\end{abstract}
			
			
			\end{frontmatter}     			
			%\linenumbers
				\thispagestyle{fancy}
		
			\section{Introduction}
			\label{sect:intro} 
			Spectral Clustering SC\citep{ng2002spectral, zelnik2004self} has gained a lot of importance in the recent times owing to its wide applicability. Some of its applications include classification, grouping and segmentation\citep{shi2000normalized}. SC is a simple method as it requires only pairwise similarity among data points. The method is data driven and easy to implement, thus, making it suitable for a variety of applications.
			\subsection{Motivation}
			SC  overcomes the 	challenges faced by traditional clustering techniques such as clustering non-convex data, and does not make any strong assumptions on the structure	of the data.
			%	From the schematic diagram in Fig. \ref{fig:schematic}, 	it is seen that 
			Construction of affinity matrix is a key step in SC. In order to enhance the SC technique, several variations to affinity matrix construction have been proposed \citep{zhang2011local,yang2011spectral}. For the sake of brevity, we discussed a few of these in the following section. We observed that the local properties play an important role in defining pairwise similarity(or affinity). Taking this into consideration, we used Topological Node Features(TNF)\citep{dahm2015efficient} to capture local characteristics and enhance the construction of affinity matrix.\\
			\textbf{Our Contribution}\\
			%\begin{enumerate}
			1. A proposed generic framework which accounts for local characteristics such as local density, spatial nearness, and structural similarity. This framework can be adapted to data of different characteristics.\\
			2. The proposed technique uses clustering coefficient TNF as local density feature in the affinity metric.\\
			3. Local structure is captured by the Summation Index(SI) TNF.\\
%		\end{enumerate}		
	%	To the best of our knowledge, this is the first method to use TNFs to enhance the affinity matrix construction in SC. 		
			The outline of this paper is as follows: Section 2, explains the state-of-the-art methods observed in literature. Section 3, briefly presents the traditional SC algorithm as given by \cite{ng2002spectral}. Section 4, describes the related theory and modeling of data. Section 5, explains the proposed TNF based framework.	Section 6 discusses the algorithm for proposed affinity matrix creation. The discussions on the results obtained in comparison with standard techniques in SC are presented in Section 7. Section 8 describes the conclusions and suggests possible future extensions.
%				\begin{figure}
%					\includegraphics[scale=.2]{figures/schematic.png}
%					\caption[Summation Index]{Schematic diagram of 
%						Spectral clustering.} 
%					\label{fig:schematic}
%				\end{figure}
			\section{Related Work}
			The following is a quick review of the recent methods proposed for the construction of effective affinity matrices. Typical similarity between points $p_i$, $p_j$ is calculated using Gaussian kernel function.
			\begin{equation}
			\hat{A}_{ij } = exp(\frac{-|| p_i- p_j ||^2}{2\sigma^{2}})
			\end{equation}
			Where %			$d(p_i,p_j)$ is the distance between $p_i$ and $p_j$ and 
			 $\sigma$ is the Gaussian kernel width. Estimation of the parameter $\sigma$ for a given dataset is an important problem in literature\citep{zhang2010spectral,gu2009improved}.\\ 
			 Global scaling is found to be inefficient when data comprises of different scales. \cite{zelnik2004self} have proposed self tuning SC which uses local scale parameter instead of global scale parameter.\\ 
			\cite{zhang2011local} have proposed an affinity measure based on Common Nearest Neighbors(CNN). The `similarity' noted in their work:
			\begin{flushleft}		
			\begin{equation}
				S_L(x_i,x_j)=
				\begin{cases}
					exp(\frac{-d(x_i,x_j)^2}{2\sigma^2(CNN(x_i,x_j)+1)})
					 &  i \neq j \\
				%	exp(-d(x_i,x_j)^2/(2\sigma^2(CNN(x_i,x_j)+1))) 
				%&  i \neq j \\
					0 											   
					&	i=j
				\end{cases}
			\end{equation}
			\end{flushleft}
			where $x_i,x_j  \in$ P, the set of all data points.	$\sigma$ is the 
			Gaussian scale parameter and CNN($x_i,x_j$) is the 
			number of common nearest 
			neighbors between $x_i,x_j$.\\
			\citep{yang2011spectral} have proposed a 
			density-based 
			similarity metric for efficient affinity matrix construction. 
			According to their method, if two points in a graph are 
			connected 
			by a path, which goes through a high density region, then 
			they 
			are said to be more similar.\\
			%An evolutionary method to arrive at the optimal affinity matrix was proposed by Chrysouli and Tefas\citep{chrysouli2015spectral}. In this method, an initial population of affinity matrices is considered. After fixing appropriate clustering criteria, genetic operations are performed on the initial population to arrive 			at the optimal affinity matrix. This method is sensitive to the initial population(\citep{chrysouli2015spectral}).\\
			Diao et al.\citep{diao2015spectral} have proposed a concept of 
			local projection neighborhood as a spatial area among data 
			points, where using local projection neighborhood, the authors 			defined local spatial structure based similarity.\\ 
			Beauchemin\citep{beauchemin2015density} has proposed a method to construct the affinity matrix employing a k-means based density estimator with subbagging procedure. % He based the technique on 	the theoretical work of Wong\citep{wong1980asymptotic}, in which asymptotic 	properties of k-means are used for density estimation. In his work subbagging procedure with k-means has been used for better density estimation, which has been further used in the similarity matrix construction.\\ 
			Yang et al.\citep{yang2013kernel} have proposed a fuzzy 
			distance based affinity matrix construction.\\ 
			From the above discussion we see that local information 
			plays an important role in enhancing affinity matrix 
			construction.\\
		    To this end, we have looked at the literature pertaining to TNF for capturing local information.\\
			Cordella et al.\citep{cordella2004sub} have used a simple TNF, the degree of a vertex, for identifying a subgraph isomorphism. TNFs have been used in the literature(\citep{sorlin2008parametric})
			to solve the subgraph isomorphism problem as they capture the local structure in the data effectively. Dahm et al.\citep{dahm2015efficient} have used TNF for subgraph isomorphism. 
			From literature we see that, TNFs were successfully used in capturing the local structural information. Hence, using the TNFs of the nodes in a graph, we proposed novel affinity matrix. We used work of Dahm et al.\citep{dahm2015efficient} for exploring the TNFs of the given data.\\ 
			We obtained encouraging results on shape datasets, UCI real datasets and MNIST handwriting dataset with our approach, where we incorporated the characteristics of data such as local density, spatial similarity, and structural similarity into the affinity matrix.	
			\section{SC Algorithm}
			\label{SpectralClusteringAlgorithm}
			We used the traditional SC, given by Ng et al.\citep{ng2002spectral} for our study. The steps in SC could be summarized as follows:
			\textit{\begin{enumerate}[1.]
				\item From the data points, Gaussian weighted distance is captured by the affinity matrix A.
				\item From A, a normalized Laplacian matrix L is constructed.
				\item Top k eigenvectors of L (k is the number of clusters) are computed. These vectors are further placed as columns, and rows of such matrix represent the original data points.
				\item Rows of the eigen vectors are clustered using the K-means algorithm.
				\item Original points are labeled based on results 	of the K-means clustering.
			\end{enumerate}}
			\section{Related theory}
			\label{Noveltnf}
%			Traditionally affinity metric is defined as the Gaussian distance 		between two points $p_i, p_j $ as given in Eq. (\ref{sect:intro})
			Our main contribution is a novel affinity metric which captures local characteristics effectively.
			This is accomplished with the help of TNFs.	The TNFs are essentially defined as topological information as viewed from any particular node of a graph. They are scale and rotation 	invariant.
			\subsection{Modeling of data}
			\label{graphG}
			Data points are modeled as nodes of graph G.
			A node $ p $ in G is connected to all nodes which are at a distance 
			less than or equal to $\epsilon$. The sparsity of graph is 
			controlled using the $\epsilon$ parameter.
			All points which are connected to node $p$ directly, 
			form the first neighborhood points, denoted as $\aleph(p)$.\\ 
			In the following section, we provide a framework based on 
			TNFs to estimate local features, and use them to enhance affinity 
			matrix construction.
			\section{TNF based Framework}
			TNFs calculated at each node are: node degree $ d $, clustering 
			coefficient $\phi$, and Summation Index $ \mathit{SI} $.
			\begin{enumerate}
			\item  `$ d $' for node $ p $ is given by the cardinality of 
			$\aleph(p)$.
			\item $\phi_p$ denotes the number of nodes in $\aleph(p)$ which are 
			connected among themselves. 
			Thus $\phi_p$	gives an intuitive understanding of local density at $p$. 
			\item $ \mathit{SI} $ is a way of propagating TNFs through the graph. 
			Thus it gives the power to encode neighboring structural 
			characteristics.
			\end{enumerate}			
			\begin{figure}[!h]
				\centering
				\includegraphics[scale=.2]{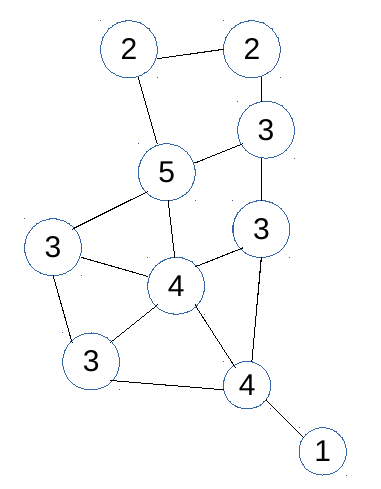}
%				\caption{Initial TNF values}
				\includegraphics[scale=.2]{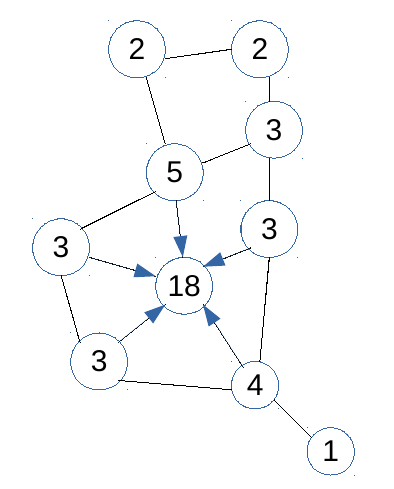}
	%			\caption{TNF values after one iteration}
				\caption[Summation Index]{(a) Initial TNF values    (b) Iteration 1 of $ \mathit{SI} $.} 
				\label{fig:sub_figure1}
			\end{figure}
			Dahm et al.\citep{dahm2015efficient} define this index as a sum of TNF values of adjacent or neighboring nodes.
			\begin{equation}
			\label{SI}
			\mathit{SI}_i(v) = 
			\begin{cases}
			feature(v) & if~ i=0 \\
			\sum\limits_{u} \mathit{SI}_{i-1}(u) & otherwise
			\end{cases}
			\end{equation}	
			where u is the node adjacent to v, $feature(v)$ is initial 
			TNF, $ d $ of a node. Fig. \ref{fig:sub_figure1} shows the evaluation of $ \mathit{SI}_2$ from $ \mathit{SI}_1$ in one iteration.
			 For every node $ p $ in G, we calculated two iterations of $ \mathit{SI} $ and placed them in $ \mathit{SI} $ vector 
			$\mathscr{V}(p)$ = 
			($ \mathit{SI}_1$, 
			$ \mathit{SI}_2$, $ \mathit{SI}_3$). This captures various levels of local structural information. \\
			We defined
			\begin{equation}
			 \zeta_{ij}=||\mathscr{V}(p_i)-\mathscr{V}(p_j)||
			 \label{zeta}
			\end{equation}
			
			\subsection{Generalized Framework for Affinity definition}
			In order to enhance the affinity between any 
			two data points $p_i$, $p_j$, we propose the following 
			generalized framework:
			\begin{equation}
			 A_{i j }' = A_{ i j } * K^1_ {i j} * K^2_ {i j} * K^3_ {i j}	
			 \label{genframe}
			\end{equation}	
			where  $A_{i j }$ is the traditional affinity, $K^1(.)$ represents density information. 
			$ K^2(.) $ represents local similarity in terms of spatial nearness. $K^3(.)$ represents local structural similarity, and $A_{i j }'$ is defined as product
			%hadamard product\citep{r1974hadamard} 
			of these individual kernels.\\
			Multiple local features are incorporated using various kernels. In this method there is a risk of over-fitting. Unnecessary information might lead to ineffective affinity metric definition as shown in some of the results in Sec \ref{results}. According to the dataset considered, appropriate local features have to be incorporated into the generalized 
			definition of metric.
			\section{Proposed Affinity matrix creation}
			The steps in affinity matrix creation employed in our method 
			are:
			\textit{
			\begin{enumerate}[1.]
				\item Model the datapoints as a graph G as explained in Sec \ref{graphG}.
				\item Let 	$\tau_{ij} = D( p_i, p_j )$ denote any standard distance(eg. Euclidean) defined over the given data points. 
				\item At each node p, calculate the following TNFs:
				\begin{enumerate}[a.]
					\item Degree of node ($ d_p $)
					\item Clustering coefficient ($\phi_p$)
					\item SI vector $\mathscr{V}_p$=($ \mathit{{SI}_1} $, $ 
					\mathit{{SI}_2} $, $\mathit{{SI}_3} $)
				\end{enumerate}
				\item We defined the similarity $ A_{ij}$ 
				between any two nodes 
				$ p_i, p_j $ as:
					\begin{equation}
					A_{ij} = \beta_{ij} * 
				\bigg( 1+\frac{1}{1+log(1+\zeta_{ij})}  \bigg) \vspace{.07cm}
					\label{aff3}
					\end{equation}						
				where			\begin{equation}
					\beta_{ij} =exp\bigg(\frac{-\tau_{ij}^2*\delta_{ij}}{2\sigma^{2}} \bigg) * 	\eta_{ij}
					\label{aff2}
					\end{equation}
			where $\delta_{ij} = abs(\phi_{i}- \phi_{j}$),  $\eta_{ij}$ is the number of common points between $\aleph(p_i), \aleph(p_j)$, and $ \sigma$ is the scale parameter of the Gaussian function. 								
			\end{enumerate}} % textit done
			Elucidating the saliency features of $A_{ij}$, the expression of $A_{ij}$ captures local density, common neighbors, and Summation Indices in the following way.\\
	%		$\alpha_{ij}$ is the density based affinity between nodes $p_i, p_j$. 
	 	%	In the definition of affinity $ A_{ij}$ in Eq.  \ref{aff3} we see that there are two terms.  which uses the structural similarity between points. \\
	 		In Eq.  (\ref{aff2}), the expression for $\beta_{ij}$  incorporates spatial nearness in the form of $\eta_{ij}$. 
			We also note that the exponential term of $\beta_{ij}$ involves the traditional distance $\tau_{ij}$ scaled with  $\delta_{ij}$. 
			Thus for points with similar density, the effective affinity will be pronounced. \\
			 %$\delta_{ij}$ is less if two points are from neighborhoods of similar density($\phi_{ij}$). 
			%If $\delta_{ij}$ is less, the distance between them is further reduced, increasing the similarity between nodes. If the $\delta_{ij}$ is large and distance is enhanced and similarity is reduced.\\
		%	In Eq.  \ref{aff2},	$\alpha_{ij}$ is scaled with $\eta_{ij}$. 
	%		If $\eta_{ij}$ is high, then it signifies that their $\epsilon$ neighborhoods are overlapping.  This gives a measure of nearness of two nodes in G and increases their probability of them belonging to same cluster.\\ 
			The second term in Eq.  (\ref{aff3}) has $\zeta_{ij}$ as the argument of log function  in the denominator.
			 Since $\zeta_{ij}$ is the difference between the local structural information  of $p_i,p_j$, the affinity increases with decrease in $\zeta_{ij}$.\\
%			  If the local structures of two points $p_i,p_j$ are similar, then the $\zeta_{ij}$ will be less. If local structures are very dissimilar, then the $\zeta_{ij}$ will be high. Affinity and $\zeta$ are inversely related.\\
			Thus in the proposed affinity measure $A_{ij}$, we are able to strengthen or penalize the traditional affinity according to local topological graph properties.  This enables our method to perform better across different types of datasets.
			%			We have organized our contributions in two formats. Affinity given by Eq.  \ref{aff2} is denoted as TNF. Affinity given by Eq.  \ref{aff3} is denoted as TNF2.			
			%So in one measure we have spatial information, structural/graphical  information and density information. 
			\subsection{Effectiveness of TNFs}
			\begin{figure}[!h]
				\centering
				\includegraphics[scale=.12]{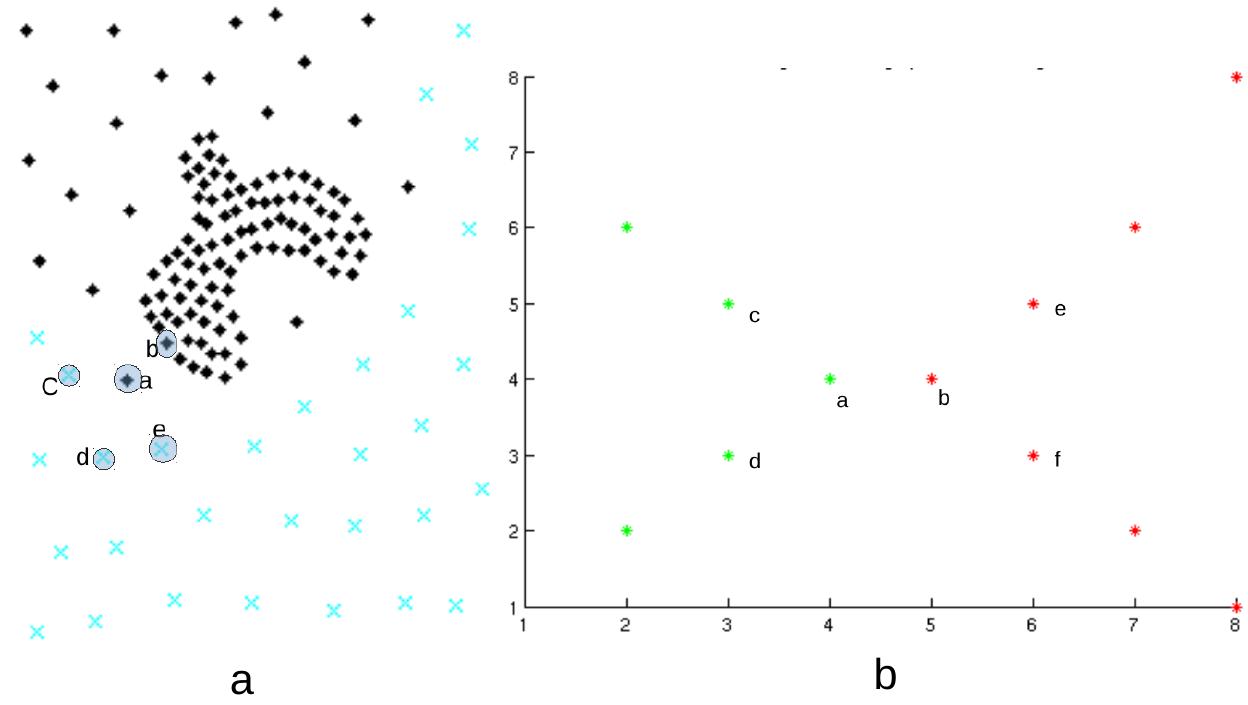}
				\caption{(a) Dataset 1 (b) Dataset 2 }
				\label{compare_figure}
			\end{figure}
\begin{figure}[!h]
				\includegraphics[width=\textwidth/2]{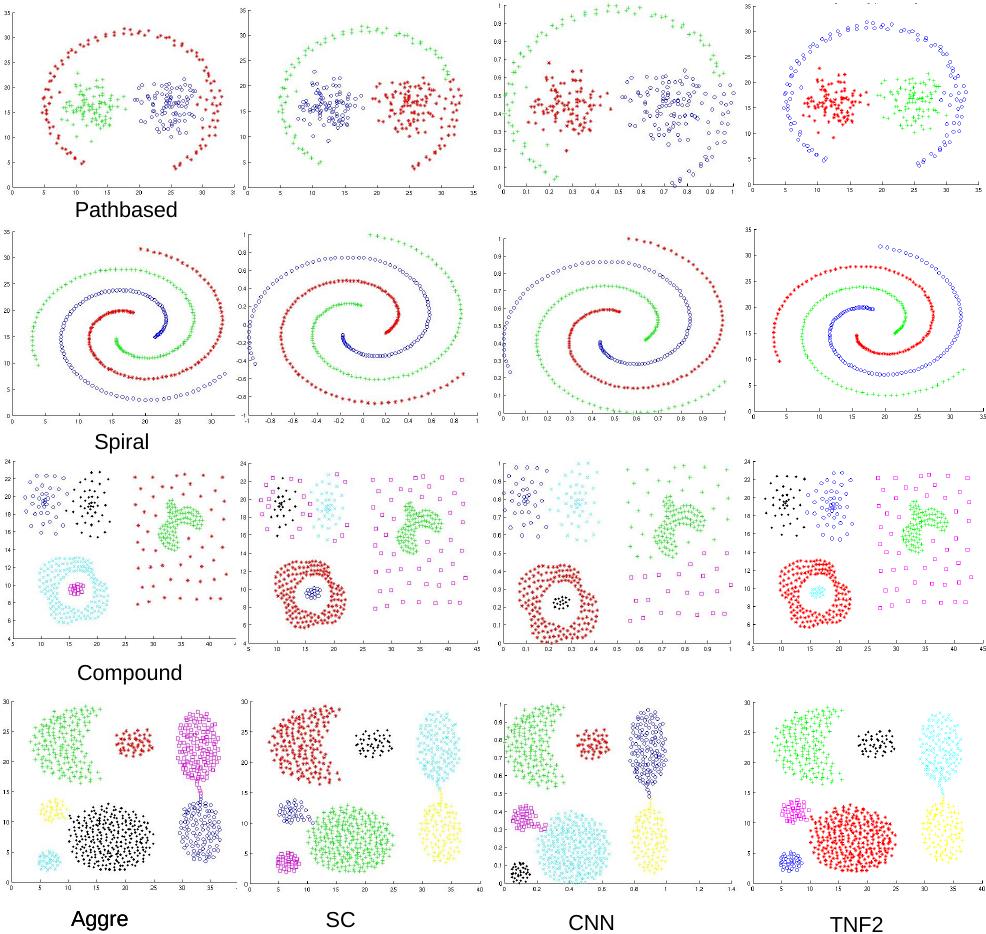}
				\caption[Results of SC, CNN,  and TNF2]{First column shows 	original datasets, second column shows results of NJW. Third column is the result of CNN based method. Fourth column is the result of the proposed algorithm} 
				\label{fig:subfigureExample1}
			\end{figure}

			\begin{table}[!h]
			\caption{Comparison of methods using ARI metric on 	Shape datasets}
					\label{table:11}
					\centering	\begin{tabular}{|p{1cm}|p{1.5cm}|p{1.3cm}|p{1.3cm}|p{1.25cm}|}
						\hline
						Affinity & Aff1 & TNF1 & TNF2\\
						\hline					
						A(a,b) & 9.66e-92  & 3.71e-32 & 3.99e-32\\ %					\hline	
						A(a,c) & 6.65e-158 & 2.38e-44 & 2.56e-44\\
						A(a,d) & 3.85e-138 & 3.09e-39 & 3.33e-39 \\
						A(a,e) & 5.88e-142 & 2.75e-28 & 2.97e-28\\
						\hline
					\end{tabular} 				
			\end{table}
			As part of our first experiment, we considered a part of Compound dataset\citep{Lichman:2013} shown in Fig.  \ref{compare_figure}(a), to highlight the working of our method. Consider points `a', `b', `c', `d', `e' from the figure. NJW\citep{ng2002spectral} wrongly assigns point `a' into the cluster in the center whereas our technique classifies it correctly( Fig.   \ref{fig:subfigureExample1}). \\
			%With further analysis we see t
			The various types of affinities  between `a' and surrounding points `b',`c',`d',`e'  are shown in Table \ref{table:11}. Aff1 refers to the Gaussian kernel distance( $\hat{A}_{ij }$). TNF1 is the affinity proposed in Eq. (\ref{aff2}), which includes local density and common neighbor parameters. TNF2 refers to the affinity proposed in Eq. (\ref{aff3}), which includes structural properties along with density and common neighbor properties.\\ 
			From the Table \ref{table:11}, we see that in the case of Aff1:
			A(a,b) $>$ A(a,d) $>$ A(a,e) $>$ A(a,c). This led to wrong clustering of point `a'. Whereas in case of affinity TNF2 : A(a,e) $>$ A(a,b) $>$ A(a,d) $>$ A(a,c). This led to a correct clustering of point `a'.\\
			The second experiment we conducted is on data given in Fig.  \ref{compare_figure}(b). The affinities between points `a', `b' are listed in Table \ref{table:12}. From the table we can see that the Aff1 between points is same but the values of TNF2 between points is different. This is to show that even when the Gaussian kernel distance between points does not show variation, structural properties can differentiate between points.\\
			This shows that our method which incorporates density and structural properties will lead to effective similarity between points.
%			 \begin{figure}
%					\includegraphics[scale=.1]{jpeg/lines_2.png}
%					\caption{Comparison of affinities over few local points}
%					\label{lines_figure}
%			\end{figure}
		\begin{table}
		\caption{Comparison of methods using ARI metric on 	Shape datasets}
		\label{table:12}
			\centering	\begin{tabular}{|p{1cm}|p{1cm}|p{1cm}|p{1cm}|p{1cm}|p{1cm}|}
				\hline
				Affinity & A(a,b) & A(a,c) & A(a,d) & A(b,e) & A(b,f) \\
				\hline					
				Aff1 & 8.2 &  8.2 & 8.2 & 8.2 & 8.2  \\ %1.4765 \\ %					\hline	
				TNF2 & .6402 & .2262 & .2262 & 1448 & .2260 \\ %1.1881 \\
%				A(a,d) & 8.2 & .2262 \\ %1.1881 \\
%				A(b,e) & 8.2 & .1448 \\ %1.1806 \\
%				A(b,f) & 8.2 & .2260 \\ %1.1867 \\
				\hline
			\end{tabular} 		
		\end{table}	
			\section{Results and Analysis}
			\label{results}
			In this section, we demonstrate the results of proposed method applied on three different types of datasets.
			The comparative results with respect to the state of the art existing techniques demonstrate the effectiveness of our method.\\			
			For experimentation, from Eq.(\ref{aff3}) we  considered two cases:\\
%			\begin{enumerate}
				Case 1(TNF1): 	$A_{ij} = \beta_{ij}$\\ Here we retained only the first term which accounts for local density and spatial nearness in the data.\\
				Case 2(TNF2): $A_{ij}$, as defined in Eq. (\ref{aff3}), which incorporates structural information in addition to $\beta_{ij}$.\\
%			\end{enumerate}
			We observed that the structural similarity term plays an important  role in some cases. For example in the case of Wine dataset (Table \ref{table:4}), by including structural similarity, we obtained clear improvement over TNF1. Whereas in case of Glass, Iris, etc. the improvement is not significant. \\	
			However compared to other methods, SC by NJW\citep{ng2002spectral}, and self-tuning(ST) SC proposed by Perona and Manor\citep{zelnik2004self}, and  Common nearest neighbors based method given by Zhang et al.\citep{zhang2011local} both TNF1 and TNF2 have done well. We considered Self Tuning with local scaling\citep{zelnik2004self}, which in general performs better than the other variation proposed by the same authors. \\
			In our experiments we used three types of metrics for comparison: Adjusted Rand Index(ARI)\citep{rand1971objective}, Normalized Mutual Information(NMI)\citep{strehl2003cluster}, Clustering Error(CE)\citep{jordan2004learning}. The values of NMI and ARI approach unity as the result goes closer to the ground truth. The metric CE represents the error in clustering that tends to null as the clustering accuracy increases.
			\subsection{Shape datasets}
			In the 2D shape datasets\citep{Lichman:2013}, we considered six examples for our experiments namely, Compound, Aggre, Flame, Jain, Pathbased, and Spiral.	The datasets present challenges such as varying density, connectedness of data etc. Some of the sample results are displayed in Table \ref{table:1}.\\ % and Fig. \ref{fig:subfigureExample1}. \\
		%	Our goal is to formulate a method which can handle this variation of characteristics and give us optimal clusters. In order to effectively cluster the data, the algorithm should take into consideration the 
		%From the results obtained by the three methods in synthetic shape datasets displayed in, we can see that our method performs better than traditional SC, STLS and CNN.  Our method is able to overcome the challenges of these datasets and give optimal clusters.\\
			In the current set of experiments,  the $\sigma$ value is chosen empirically. We experimented with $\sigma$	varying from .01 to 10 with an interval of .01. 
			%After examining the results we were able to pin point an exact $\sigma$ value which gives optimal result.
			 Selection of optimal sigma for spectral clustering is an open problem and a few methods have been proposed in the literature\citep{zhang2010spectral,gu2009improved}. We note that in all cases both TNF1 and TNF2 are performing better. TNF2 does not show significant improvement over TNF1.
		%	\squeezeup
			\begin{table}[!h]
		\caption{Comparison of methods using ARI metric on 	Shape datasets}
		\label{table:1}
				\centering	\begin{tabular}{|p{1cm}|p{.8cm}|p{.8cm}|p{1cm}|p{.8cm}|p{.8cm}|p{.8cm}|}
					\hline
					&  &  & Datasets & & &\\
					\hline					
					Method & Comp & Aggre & Flame & Jain & Path & Spiral\\
					\hline	
					NCUTS & 0.9405 & 0.9869 & 1 & 1 & 0.7143 & 1\\
					ST         & 0.5184 & 0.9642 & 0.625 & 0.9444 & 0.5138 & 0.0781\\
					CNN     & 0.8955 & 0.9833 &	0.9667 & 1 & 0.7187 & 1\\
					TNF1   & 0.9972 & 1 & 1 & 1 & 0.9899 & 1\\
					TNF2   & \textbf{0.9972} & \textbf{1} & \textbf{1} 
					& \textbf{1} & 	\textbf{1} & \textbf{1} \\
					\hline
				\end{tabular} 			
			\end{table}
		%	\squeezeup
			\begin{table}[!h]
\caption{Comparison of methods using NMI metric on Shape datasets}
			\label{table:2}
				\centering
					\begin{tabular}{|p{1.05cm}|p{.8cm}|p{.8cm}|p{1cm}|p{.8cm}|p{.8cm}|p{.8cm}|}
					\hline
					&  &  & Datasets & & &\\
					\hline					
					Method & Comp & Aggre & Flame & Jain & Path & 	Spiral\\
					\hline		
					NCUTS &	0.9171&	0.9824&	1 &	1	& 	0.7825&	1\\
					ST &0.7632&	0.9661&	0.564 &	0.8961&	0.5869 &	0.1716\\
					CNN & .9120 & 0.9808 & 0.9269 & 1 & 0.7728 & 1 \\
					TNF1 & 0.9924 & 1 & 1 & 1 & 0.9829 & 1\\
					TNF2 & \textbf{0.9924} & \textbf{1} & \textbf{1} & \textbf{1} & \textbf{1} & \textbf{1}\\			
					\hline
				\end{tabular}
			\end{table} 
			\begin{table}[!h] 
			\caption{Comparison of methods using CE metric on Shape datasets}
			\label{table:3}
			\centering
				\begin{tabular}{|p{1cm}|p{.8cm}|p{.8cm}|p{1cm}|p{.8cm}|p{.8cm}|p{.8cm}|}
					\hline
					&  &  & Datasets & & &\\
					\hline					
					Method & Comp & Aggre & Flame & Jain & Path & Spiral\\
					\hline	
					NCUTS &	0.0526&	0.0063&	0&	0&	0.1133&	0\\
					ST &0.3559&	0.0165&	0.1042&	0.0134&	0.2133&0.4808\\
					CNN & 0.0702 & 0.0076 & 0.0083 & 0 & 0.1100 & 0\\
					TNF1 &	0.0025 & 0 & 0 & 0 & 0.0033 & 0\\
					TNF2 & \textbf{0.0025} & \textbf{0} & \textbf{0} &	
					\textbf{0} & \textbf{0} & \textbf{0}\\			
					\hline
				\end{tabular}
			\end{table}

			\subsection{Real datasets}
			We considered UCI real datasets(\citep{Lichman:2013}) as second type of dataset. 
			%The details of the five datasets, we considered for our  			experiment are given in Table \ref{table_prop}. 
			These datasets are collected from real scenarios and have varied number of features and distributions.
%			\begin{table}
%				\title{Table1}
%				\begin{tabular}{ |p{2.2cm}| p{.8cm}| p{.8cm}| p{.8cm}| p{.8cm}| p{.8cm}|}	\hline
%					Dataset & Wine & Glass & Iris & Ion & Sonar\\		\hline 
%					No of instances & 178 & 214 & 150 & 351 & 	208\\  
%					No of attributes & 13 & 9 & 4 & 34 & 60\\
%					No of clusters & 3 & 6 & 3 & 2 & 2\\%					\hline
%				\end{tabular}
%				\caption{Attributes of real UCI datasets}
%				\label{table_prop}
%			\end{table}
%			\squeezeup
			Results of TNF1, TNF2 in comparison with other SC methods are given in Tables \ref{table:4}, \ref{table:5}, \ref{table:6}. From the results shown in Table \ref{table:4}, we see that TNF2 shows improvement over TNF1 in Wine, Glass and Iris datasets. In case of Ion dataset, the result remains same. In case of Sonar dataset, TNF1 is better than TNF2 with respect to ARI metric. The structure of the dataset then determines which TNFs help in creating effective affinity matrix.
			\begin{table}[!h]
				\centering
			\sq	\caption{Comparison of methods using ARI metric on UCI datasets}
					\label{table:4}
				\begin{tabular}{ |p{1cm}| p{1cm}| p{1cm}| p{1cm}| p{1cm}| p{1cm}|}
					\hline
					&  &  & Datasets & & \\
					\hline
					Methods & Wine & Glass & Iris & Ion & Sonar\\
					\hline 
					NCUTS & 0.4127 & 0.2876 & 0.8161 & 0.6647 & 0.0630\\
					ST & 0.319 & 0.2352 & 0.7580 & 0.2184 & 0\\
					CNN & 0.9149 & 0.2806 & 0.7592 & 0.6926 & 	0.0289\\
					TNF1 & 0.7782 & 0.3559 & 0.8683 & 0.7020 & \textbf{0.1438}\\
					TNF2 & \textbf{0.9471} & \textbf{0.3575} & \textbf{0.8858} & \textbf{0.7020} & 0.1224\\
					\hline
				\end{tabular}			
			\end{table}		%	\squeezeup
			\begin{table}[!h]
				\sq	\caption{Comparison of methods using NMI metric on 	UCI datasets}
			\label{table:5}
				\centering
				\begin{tabular}{ |p{1cm}| p{1cm}| p{1cm}| p{1.1cm}| p{1cm}| p{1cm}|}
					\hline
					&  &  & Datasets & & \\
					\hline
					Methods & Wine & Glass & Iris & Ion & Sonar\\
					\hline 
					SC & 0.4554 & 0.4670 & 0.8058 & 0.5463 & 0.0995\\
					ST & 0.395 & 0.4143 & 0.7856 & 0.2214 & 0.0030\\ 
					CNN & 0.8926 & 0.4406 &  0.8058 & 0.5820 & 0.0615\\ 
					TNF1 & 0.7696 & 0.4943 & 0.8572 & 0.6116 & 0.1757\\
					TNF2 & \textbf{0.9276} & \textbf{0.5035} &\textbf{ 0.8705} & \textbf{0.6116} & \textbf{0.1946}\\
					\hline
				\end{tabular}			
			\end{table}%			\squeezeup
			\begin{table}[!h]
	\sq		\caption{Comparison of methods using CE metric on UCI datasets}
			\label{table:6}		
			\centering
					\begin{tabular}{ |p{1cm}| p{1cm}| p{1cm}| p{1cm}| p{1cm}| p{1cm}|}
					\hline		 &  &  & Datasets & & \\
					\hline
					Methods & Wine & Glass & Iris & Ion & Sonar\\
					\hline 
					SC & 0.2809 & 0.4393 & 0.0667 & 0.0912 & 0.3702\\
					ST & 0.4440 & 0.5373 & 0.0930 & 0.2650 & 0.4760\\
					CNN & 0.0281 & 0.4533 & 0.0933 & 0.0826 & 0.4087\\
					TNF1 & 0 & 0.3645 & 0.0467 & 0.0798 & 0.3077\\
					TNF2  & \textbf{0} & \textbf{0.3598} & \textbf{0} & \textbf{0.0800} &\textbf{0.3221}\\
					\hline
				\end{tabular}			
			\end{table}
			\begin{table}[!h]
	\sq		\caption{Comparison of methods using ARI metric on 	MNSIT datasets}	
			\label{table:7}
				\centering
				\begin{tabular}{ |p{1.5cm}| p{1.5cm}| p{1.5cm}| p{1.5cm}|}
					\hline
					&  & Datasets & \\
					\hline
					Methods & \{0,8\} & \{3,5,8\} & \{1,2,3,4\} \\
					\hline 
					SC & 1 & 0.5657 & 0.3740\\
					ST & 1 & 0.4535 & 0.2297\\
					CNN & 1 & 0.5682 & 0.33102\\
					TNF1 & \textbf{1} & \textbf{0.8159} & \textbf{0.6340}\\
					\hline
				\end{tabular}			
			\end{table}	
			\begin{table}[!h]
\caption{Comparison of methods using NMI metric on MNSIT datasets}
			\label{table:8}								
				\centering
				\begin{tabular}{ |p{1.5cm}| p{1.5cm}| p{1.5cm}| p{1.5cm}|}
					\hline 
					&  & Datasets & \\
					\hline
					Methods & \{0,8\} & \{3,5,8\} & \{1,2,3,4\} \\
					\hline 
					SC & 1 & 0.7502 & 0.6216\\
					ST & 1 & 0.6570 & 0.5221\\
					CNN & 1 & 0.7545 & 0.6325\\
				%	TNF1 & 1 & 0.5682 & 0.3295\\
					TNF1 & 1 & \textbf{0.7802} & \textbf{0.6835}\\
					\hline
				\end{tabular}
			\end{table}	
			\begin{table}[!h]
	\caption{Comparison of methods using CE metric on MNSIT datasets} 
					\label{table:9}
				\centering
				\begin{tabular}{ |p{1.5cm}| p{1.5cm}| p{1.5cm}| p{1.5cm}|}
					\hline
					&  & Datasets & \\
					\hline
					Methods & \{0,8\} & \{3,5,8\} & \{1,2,3,4\} \\
					\hline 
					SC & 0 & 0.3367 & 0.4050\\
					ST & 0 & 0.4533 & 0.6650\\
					CNN & 0 & 0.3350 & 0.5013\\
					%TNF1 & 0 & 0.3400 & 0.5025\\
					TNF1 & \textbf{0} & \textbf{.0667} & 	\textbf{.1800}\\
					\hline
				\end{tabular}
			\end{table}
			\subsection{Handwritten datasets}
			MNIST dataset given by lecun et al.\citep{lecun1998gradient} is a handwritten digits database. It has a training set of 60,000 examples and test set of 10,000 samples. For each of the ten digits, there is a test set of 1000 samples. All the samples are images of size 28x28.\\ 
			For our experiments, we considered 200 samples of each digit. We tested our method on some of challenging test cases such as \{0,8\}, \{3,5,8\}, \{1,2,3,4\}. We employed TNF1 for this dataset. Tables \ref{table:7}, \ref{table:8}, \ref{table:9} summarize the results that again reiterate the greater efficacy of our technique.
			\section{Conclusion}
			Traditionally, in a SC algorithm, the pairwise similarity between data points is estimated using a Gaussian kernel function. In this work, we proposed a novel similarity measure based on local properties. Properties, such as local neighborhood, local density information, and local structure were estimated using TNFs and were incorporated into the construction of pairwise affinity. Using topological graph properties, we were able to enhance or penalize the pairwise similarity. Our experiments on synthetic, real and handwriting datasets show that proposed TNF based technique improved the effectiveness of SC. In our future work, we would like  to adapt this framework for different applications such as Image segmentation etc. The framework can also be strengthened by  assimilating more topological node features such as Listing index, Tree index \citep{dahm2015efficient}. 
		\section*{Acknowledgments}
		We dedicate our work to the founder chancellor of Sri Sathya Sai Institute of Higher Learning, Bhagawan Sri Sathya Sai Baba.
		
			\bibliographystyle{model2-names}
		%	\bibliographystyle{IEEEtran}
		%	\bibliographystyle{PRL}
		%	\bibliography{IEEEabrv,report}
		\bibliography{report1}
		\end{document}